%% file: main.tex
\documentclass{article}

 \usepackage[preprint]{neurips_2026}

\usepackage[utf8]{inputenc} 
\usepackage[T1]{fontenc}    
\usepackage{hyperref}       
\usepackage{url}            
\usepackage{booktabs}       
\usepackage{amsfonts}       
\usepackage{nicefrac}       
\usepackage{microtype}      
\usepackage{xcolor}         

\usepackage[T1]{fontenc}
\usepackage{amsmath, amssymb, amsfonts}
\usepackage{mathtools}
\usepackage{bm}
\usepackage{booktabs}
\usepackage{graphicx}
\usepackage[nameinlink,capitalize]{cleveref}
\crefname{section}{Sec.}{Secs.}
\Crefname{section}{Section}{Sections}
\crefname{table}{Tab.}{Tabs.}
\Crefname{table}{Table}{Tables}
\crefname{appendix}{App.}{Apps.}
\Crefname{appendix}{Appendix}{Appendices}
\usepackage{enumitem}
\usepackage{parskip}
\usepackage{tikz}
\usetikzlibrary{arrows.meta, positioning, fit, calc, backgrounds}
\usepackage{caption}
\usepackage{gensymb}
\usepackage{multirow}
\hypersetup{
  colorlinks=true,
  linkcolor=blue!50!black,
  citecolor=blue!50!black,
  urlcolor=blue!50!black,
}

\usepackage{xcolor}

\input{math_commands}
\input{preamble}

\title{ArtSplat: Feed-Forward Articulated 3D Gaussian Splatting from Sparse Multi-State Uncalibrated Views}

\author{%
  \textbf{Inseo Lee}$^1$ \quad
  \textbf{Yoonji Kim}$^2$ \quad
  \textbf{Eugene Sohn}$^1$ \quad
  \textbf{Jiwoong Lee}$^1$ \\
  \textbf{Jungmin You}$^1$ \quad
  \textbf{Joonseok Lee}$^{1,\dagger}$ \quad
  \textbf{Jin-Hwa Kim}$^{1,3,\dagger}$ \\
  $^1$Seoul National University \quad
  $^2$Sogang University \quad
  $^3$NAVER AI Lab \\
  \texttt{\{ian.lee, qeugene, dlwldnd5689, ujxlan409, joonseok\}@snu.ac.kr} \\
  \texttt{wayuj@sogang.ac.kr} \quad
  \texttt{j1nhwa.kim@navercorp.com} \\ \\
  {\large \href{https://artsplat.github.io/}{\texttt{https://artsplat.github.io}}}
}

\begin{document}

\renewcommand{\thefootnote}{}
\footnotetext{$^{\dagger}$Co-corresponding authors}
\renewcommand{\thefootnote}{\arabic{footnote}}

\maketitle

\vspace{-0.8em}

\input{section/0_abstract}
\input{section/1_intro}
\input{section/2_related_work}
\input{section/3_method}
\input{section/4_experiments}
\input{section/5_conclusion}

\section*{Acknowledgments}
This work was also supported by Samsung Electronics, Youlchon Foundation, National Research Foundation of Korea (NRF) grants (RS-2021-NR05515, RS-2024-00336576, RS-2023-0022663), and the Institute for Information \& Communication Technology Planning \& Evaluation (IITP) grants (RS-2022-II220264, RS-2024-00353131) funded by the Korean government.

\medskip
{
    \small
    \bibliographystyle{abbrv}
    \bibliography{main}
}

\newpage
\input{section/X_suppl}



\end{document}

%% file: math_commands.tex
\newcommand{\R}{\mathbb{R}}

\newcommand{\Iset}{\mathcal{I}}
\newcommand{\Iview}[1]{\mathbf{I}_{#1}}     
\newcommand{\statevar}{s}
\newcommand{\stateset}{\{0,1\}}

\newcommand{\zstate}[1]{\mathbf{z}_{#1}}            
\newcommand{\zstatebb}[1]{\mathbf{z}_{#1}^{\mathrm{bb}}}    
\newcommand{\ztilde}[1]{\tilde{\mathbf{z}}_{#1}}     

\newcommand{\Fpatch}[1]{F^{(#1)}}
\newcommand{\Pmap}{\mathbf{P}}

\newcommand{\Jmap}{\mathbf{J}}
\newcommand{\Jinv}{\mathbf{J}^{\textit{inv}}}
\newcommand{\Jvar}{\mathbf{J}^{\textit{var}}}
\newcommand{\jaxis}{\mathbf{a}}
\newcommand{\jpivot}{\mathbf{p}}
\newcommand{\jtype}{\boldsymbol{\tau}}

\newcommand{\cinv}{\mathbf{c}^{\textit{inv}}}
\newcommand{\cvar}[1]{\mathbf{c}^{\textit{var}}_{#1}}

\newcommand{\Gset}{\mathcal{G}}
\newcommand{\Gauss}{g}

\newcommand{\Lloss}[1]{\mathcal{L}_{\textit{#1}}}

%% file: preamble.tex
\newif\ifinseo
\inseotrue

\newcommand{\paragrapht}[1]{\noindent\textbf{#1}}

%% file: section/0_abstract.tex
\begin{abstract}
    Articulated object reconstruction from sparse-view images is an ill-posed problem that requires simultaneous inference of geometry and underlying articulation structure.
    Existing methods for articulated object reconstruction based on NeRF and 3D Gaussian Splatting (3DGS) typically rely on dense views or strong priors (\textit{e.g.}, depth maps, joint types, predefined number of joints) and require costly per-object optimization.
    In this paper, we propose ArtSplat, the first feed-forward framework for articulated 3D Gaussian Splatting.
    It reconstructs both geometry and joint parameters from sparse multi-view images across multiple articulation states in a single forward pass. To address the challenges of single-pass articulated reconstruction, we introduce a per-pixel joint map representation that enables the integration of joint parameter estimation into the feed-forward pipeline.
    We further propose a Cross-State Attention (CSA) mechanism with state tokens, which effectively captures discrete motion across input states.
    Experiments on 68 articulated objects from PartNet-Mobility, including both single- and multi-joint configurations, demonstrate that ArtSplat achieves competitive performance in both geometry and joint estimation, while being over 400 times faster than baselines.
\end{abstract}

%% file: section/1_intro.tex
\section{Introduction}
\label{sec:intro}

An \emph{articulated object} consists of rigid parts connected by joints, allowing the parts to move relative to one another.
These joints come in two types: revolute joints, which rotate about an axis (\emph{e.g.}, a laptop lid or a cabinet door), and prismatic joints, which translate along an axis (\emph{e.g.}, a drawer).
Representing such an object, therefore, requires not only the geometry of each part but also the joint parameters that determine how those parts move.
The \emph{articulated object reconstruction} is the task of recovering both of these from observed samples.
Unlike static reconstruction, the resulting representation should be able to move along the recovered joints into any valid pose, called an \emph{articulation state}, of its parts.
This makes it directly applicable to downstream tasks such as manipulation policy learning, sim-to-real asset generation, and digital twin construction.

Early work on articulated object reconstruction adopts implicit representations such as neural radiance fields~\cite{mildenhall2021nerf}, recovering both geometry and joint parameters from multi-state observations~\cite{jiang2022ditto,liu2023paris, tseng2022cla}.
Recently, 3D Gaussian Splatting (3DGS)~\cite{kerbl20233dgs} is widely adopted to leverage its explicit primitives and faster rendering to improve reconstruction quality~\cite{guo2025articulatedgs, kim2025screwsplat, lin2025splart, liu2025artgs, shen2025gaussianart, wu2025reartgs, yu2025part^2}.
Both approaches, however, rely on test-time optimization, still requiring fitting each object from scratch over dense (typically 50+ per articulation state) multi-view captures.
These methods do not scale well to a large number of novel objects, since both the capturing and optimization steps should be repeated for each new object, requiring tens of minutes or even hours per object.

A natural alternative is to predict the reconstruction through a single feed-forward pass, which has been successfully applied to static scenes, where the scene does not change across input views, by transformer-based geometry models~\cite{wang2024dust3r, leroy2024grounding, wang2025vggt, yang2025fast3r, zhang2025flare} and feed-forward 3DGS variants~\cite{charatan2024pixelsplat, chen2024mvsplat, xu2025depthsplat, ye2025nopose, smart2024splatt3r, jiang2025anysplat}, and to dynamic scenes by their dynamic counterparts~\cite{zhang2024monst3r, wang2025continuous, chen2025easi3r}, where an ordered sequence of views with small motion between them are given.
By contrast, articulated objects do not fit either case;
the input views are \textit{not static} as they contain parts with non-trivial movements across their articulation states, but they are still \textit{unordered}.
Given a few RGB views containing multiple articulation states, the task aims to recover geometry and infer how the parts move at the same time, from limited observations.

In order to handle both the object geometry and articulation states within a single forward pass, we propose ArtSplat, a feed-forward model that predicts 3D Gaussians with joint parameters 
from sparse uncalibrated views captured at two or more articulation states.
The core of our design is a joint prediction module that recovers how parts can move across various articulation states.
Each state is paired with its own learnable state token, while a cross-state attention block allows each token to attend to the patch tokens of another state, enabling two tokens to capture the inter-state motion.
A dual-branch DPT head~\citep{ranftl2021dpt} then decodes the patch tokens, conditioned on the state tokens, into a per-pixel joint map.
One branch predicts joint type, axis, and pivot, which remain constant across states, and the other predicts rotation angle and translation distance, which change between states.

The resulting joint map encodes joint type, axis, pivot, and motion magnitude at every pixel. Combined with 3D Gaussian primitives---whose means are unprojected from a predicted depth map and whose covariance, opacity, and color come from a Gaussian head---these per-pixel parameters articulate the Gaussians into any target articulation state for rendering.

In summary, our contributions are as follows:
\begin{itemize}[leftmargin=2em, itemsep=-0.3em]
    \item We propose ArtSplat, the first feed-forward architecture that jointly predicts 3D Gaussians and joint parameters from uncalibrated RGB views, removing the per-object test-time optimization required by prior methods.
    \item We design a joint prediction module that explicitly captures discrete inter-state motion, enabling accurate joint estimation from sparse views.
    \item Extensive experiments show that our model matches or outperforms prior baselines in reconstruction and joint-estimation quality, while running \emph{400$\times$ faster} than prior works.
\end{itemize}

%% file: section/2_related_work.tex
\section{Related work}
\label{sec:related}

\paragrapht{Articulated object reconstruction.}
Early implicit methods~\cite{liu2023paris,jiang2022ditto,tseng2022cla} optimize per-object neural fields from multi-view observations across articulation states.
Subsequent 3D Gaussian Splatting (3DGS) approaches~\cite{guo2025articulatedgs,liu2025artgs,wu2025reartgs,shen2025gaussianart,yu2025part^2} have improved rendering fidelity, yet many still rely on strong priors such as depth, joint types~\cite{lin2025splart} or predefined number of joints~\cite{liu2025artgs,wu2025reartgs}.
More recent prior-free approach~\cite{kim2025screwsplat} has attempted to relax these assumptions.
Across both paradigms, however, most methods remain formulated around a single joint, handling multi-part objects through iterative or sequential processing.
Moreover, they typically require dense multi-view samples and rely on per-object optimization, incurring substantial computational overhead that limits practical applicability.

\paragrapht{Generalizable feed-forward reconstruction.}
A parallel line of work focuses on eliminating per-scene optimization via feed-forward inference, directly predicting 3D structure from input images.
For static scenes, transformer-based models such as DUSt3R~\cite{wang2024dust3r} and its successors~\cite{leroy2024grounding,wang2025vggt,yang2025fast3r,zhang2025flare} have demonstrated remarkable capabilities in recovering 3D geometry from sparse views.
Similarly, 3DGS-based feed-forward methods~\cite{charatan2024pixelsplat,chen2024mvsplat,xu2025depthsplat,ye2025nopose,smart2024splatt3r,jiang2025anysplat} enable efficient reconstruction in a single pass.
However, these methods assume that the object remains static across all input views.
For multi-state inputs of articulated objects, they cannot separate viewpoint variation from inter-state part motion, producing ghosting artifacts.
While dynamic scene models like MonST3R~\cite{zhang2024monst3r} and others~\cite{chen2025easi3r,wang2025continuous} extend this to moving geometry, they typically assume dense temporal correspondence or small, gradual displacements across video frames.
Articulated objects, in contrast, often appear in a few discrete states with large-magnitude motion, motivating a dedicated feed-forward framework tailored to such discrete, large-displacement state transitions.

\paragrapht{Feed-forward articulated reconstruction.}
Most closely related to our work are ART~\cite{li2026art} and LARM~\cite{yuan2025larm}, which have established articulated object reconstruction in a single forward pass.
ART uses a transformer with fixed-part slots to decompose objects via SDF-based volume rendering, conditioned on Plücker ray embeddings that require known camera poses. While providing structured outputs, it is limited by a predefined part count and suffers from a heavy computational bottleneck due to dense ray sampling at rendering.
LARM, on the other hand, employs a transformer decoder to synthesize novel views conditioned on camera poses and joint states.
Rendering an image thus requires a full forward pass through the transformer with the target pose and state as input, so its inference cost scales linearly with the number of desired views and states.
In contrast, our method directly predicts 3D Gaussians and joint parameters from uncalibrated RGB images in a single forward pass, enabling efficient real-time rendering of articulated objects.

%% file: section/3_method.tex
\section{Problem formulation}
\label{sec:problem}

The \textbf{articulated object reconstruction} task takes $S$ articulation states with $V$ sparse views for each as inputs, giving a total of $N=V\cdot S$ pose-free RGB images 
$\Iset = \{\Iview{i,\statevar}\in \R^{H\times W\times 3}\}_{i=1,\dots,V,\,\statevar=1,\dots,S}$
of an articulated object with one or more movable parts at resolution $H\times W$.
An articulation between two states is either \emph{revolute}, a rotation by angle $\theta$ around an axis at a pivot point (rotation requires axis location, not just direction), or \emph{prismatic}, a displacement by $d$ along an axis.

The model is expected to encode the $N$ input images and to produce predictions for multiple heads in a single forward pass, \emph{e.g.}, \emph{camera}, \emph{depth}, \emph{joint}, and \emph{Gaussian}: 
\begin{itemize}[leftmargin=2em, itemsep=-0.3em]
  \item $\boldsymbol{\pi}_{i,\statevar}\!\in\!\R^{9}$, the per-image camera parameters from the camera head, decomposed as translation $\mathbf{t}\!\in\!\R^{3}$, unit quaternion $\mathbf{q}\!\in\!\R^{4}$, and camera field-of-view $\mathbf{f}\!\in\!\R^{2}$.
  
  \item $\mathbf{D}_{i,\statevar}\!\in\!\R^{H\times W}$ and $\mathbf{C}_{i,\statevar}\!\in\!\R^{H\times W}$, the per-pixel depth and confidence maps from the depth head. Note that the depth map determines the Gaussian means and, when unprojected with the predicted camera $\boldsymbol{\pi}_{i,\statevar}$, produces a point map $\Pmap_{i,\statevar}\!\in\!\R^{H\times W\times 3}$.
  
  \item The Gaussian set $\Gset = \{\Gauss_n\}_{n=1}^{H \times W \times N}$ is constructed from depth-derived means with Gaussian attributes predicted by a Gaussian head: scale $\mathbf{s}\!\in\!\R^{3}$, rotation quaternion $\mathbf{r}\!\in\!\R^{4}$, opacity $\alpha\!\in\!\R$, and view-dependent color represented with spherical harmonics of degree $L\!=\!4$, $\mathbf{c}\!\in\!\R^{3\times(L+1)^{2}}\!=\!\R^{75}$, yielding $86$ parameters per Gaussian.

  \item $\Jmap_{i,\statevar}\!\in\!\R^{H\times W\times 11}$, the per-pixel \textit{joint map} from the joint head, with channels for the joint type logits over \emph{static}, \emph{revolute}, and \emph{prismatic} ($3$), axis direction $\jaxis$ ($3$, unit-norm), pivot location $\jpivot$ ($3$), revolute angle $\theta$ ($1$), and prismatic displacement $d$ ($1$).
  The Gaussian transformation is conditioned on the predicted joint type: \textit{revolute} pixels use rotation parameters $\{\jaxis,\jpivot,\theta\}$, \textit{prismatic} pixels use translation parameters $\{\jaxis,d\}$, and \textit{static} pixels remain unchanged.
  
\end{itemize}

As illustrated in \cref{fig:overview}, a differentiable transformation of this Gaussian set by the joint map values produces a state-conditioned representation, which can then be rendered at any target viewpoint.

\section{ArtSplat: the proposed method}
\label{sec:method}

\begin{figure}
    \centering
    \includegraphics[width=1\linewidth]{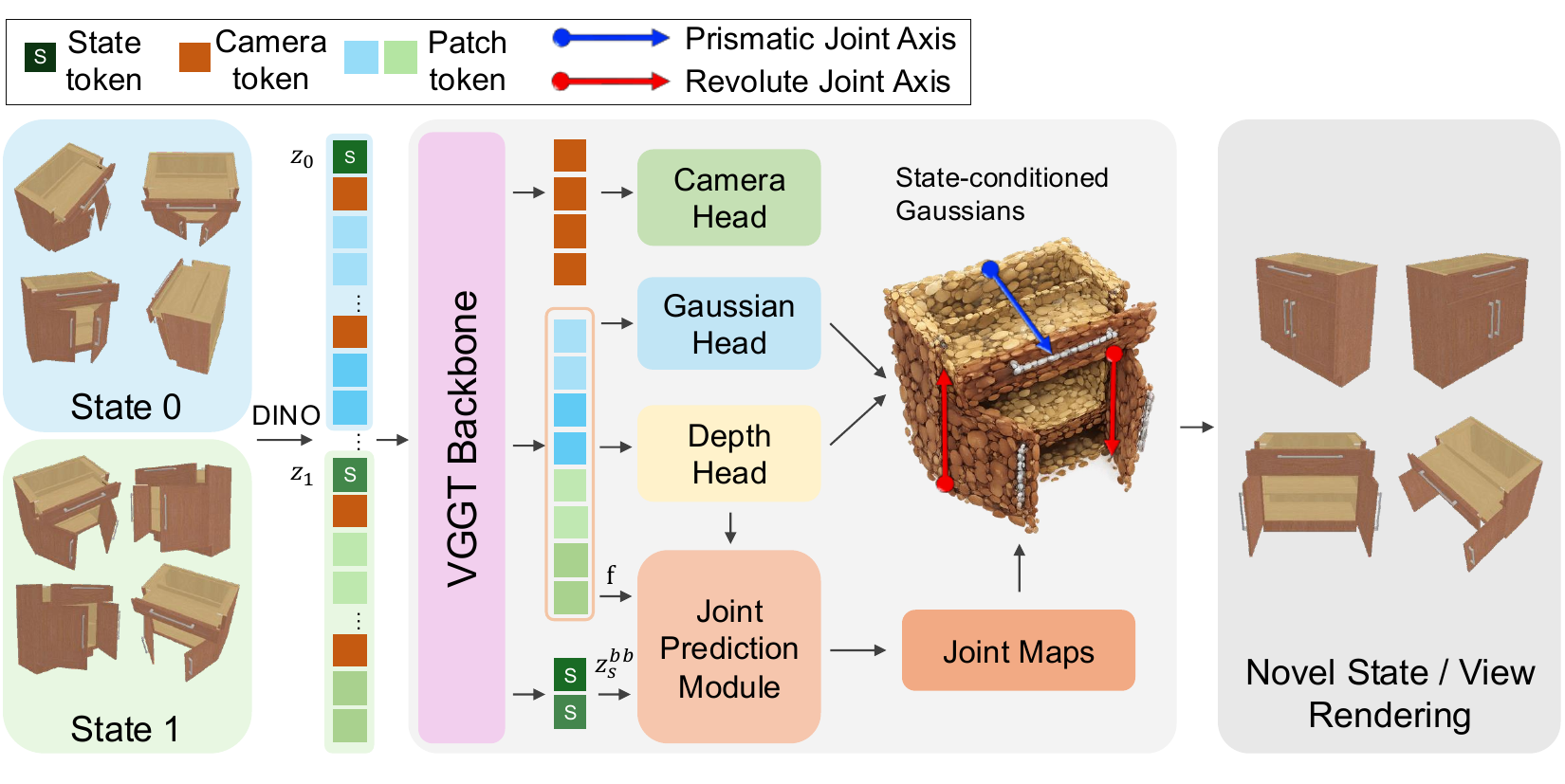}
    \caption{Overview. Given sparse multi-view images across two states, our model predicts geometry and joint parameters in a forward pass. Depth and Gaussian predictions are integrated with the joint maps to produce a state-conditioned Gaussian set, enabling articulated novel-state rendering without per-object optimization.}
    \label{fig:overview}
\end{figure}

We propose ArtSplat, a transformer-based feed-forward model that extends VGGT~\citep{wang2025vggt} from static scenes to articulated objects by introducing a per-pixel joint map representation (\cref{sec:joint_map}).
As shown in \cref{fig:overview}, ArtSplat takes sparse multi-view images of two articulation states and predicts the camera, depth, Gaussian, and joint parameters in a single forward pass.
Each input image is patchified and encoded into a set of tokens by DINOv2~\citep{oquab2023dinov2}, prepended with learnable camera and state tokens, and contextualized through $L$ layers of alternating frame and global attentions.
These outputs are decoded by task-specific heads and a joint prediction module (\cref{sec:jmpm}) into per-pixel predictions that together form a state-conditioned Gaussian set $\Gset$ (\cref{sec:scgs}).
Training is detailed in \cref{sec:training}

\subsection{Joint map representation}
\label{sec:joint_map}

To reconstruct an articulated object, a model needs to determine (i) which parts can move, (ii) along which axes, and (iii) their rotation angle or displacement.
One naive approach is to first segment the object into discrete parts and then regress one joint per part. However, the segmentation step relies on non-differentiable operations, breaking end-to-end training and making it difficult to integrate the pipeline into a feed-forward network.
Another naive approach is to reserve a fixed number of output slots, with each slot predicting the joint parameters of one part.
However, the number of parts varies across objects, and the assignment between predicted slots and ground-truth parts is often unstable during training, leading to inconsistent learning signals.

We instead represent all joint parameters as a dense, per-pixel joint map $\Jmap$, where each pixel is represented as an 11-dimensional vector defined in \cref{sec:problem}.
These per-pixel joint parameters directly apply to its Gaussian primitive, constructed on the same pixel.
By formulating articulation as a per-pixel regression, the joint head becomes fully differentiable and can be trained end-to-end with the depth and Gaussian heads, integrating articulation into our feed-forward reconstruction pipeline.

The 11 channels of the joint map $\Jmap$ can be categorized into two groups, either invariant across the articulation state or variant by pixels.
As summarized in \cref{tab:joint_map_channels},
the \emph{invariant} group $\Jinv\!\in\!\R^{9}$ contains the properties intrinsic to the joint, \emph{e.g.}, the joint type logits, the axis direction, and the pivot location.
They remain consistent for any pixel belonging to the same part across two states.
On the other hand, the \emph{variant} group $\Jvar\!\in\!\R^{2}$ contains the rotation angle~$\theta$ and translation displacement~$d$, which describe how far the part has moved \emph{at the current state} and therefore differ across states.

\subsection{Joint map prediction module}
\label{sec:jmpm}

To predict the joint map, the network must identify what changes between two states; that is, which pixels move along an axis and by how much.
We model this with three components: \textit{state tokens} that capture per-state information, a \textit{cross-state attention block} that compares the two states, and an \textit{invariant/variant DPT head} that turns the result into the per-pixel joint map.

\paragrapht{State tokens.}
We add a learnable state token per state, denoted by $\zstate{0},\zstate{1}\!\in\!\R^{D}$, inspired by the camera-token design of VGGT~\citep{wang2025vggt}.
The state token $\zstate{\statevar}$ is prepended to the $V$ views of state~$\statevar$.
As the VGGT backbone's alternating frame- and global-attention layers process the full sequence, $\zstate{\statevar}$ attends to all $V$ views of state~$\statevar$.
We denote the resulting backbone output as $\zstatebb{\statevar}\!\in\!\R^{D}$.

\paragrapht{Cross-state attention}.
To strengthen the comparison between the two states, we introduce a single Cross-State Attention (CSA) block in which $\zstatebb{\statevar}$ attends to the other state's image features. Let $\Fpatch{\statevar}\!\in\!\R^{V \times N_p \times D}$ denote the patch tokens at the backbone's last block, collected across all $V$ images of state $\statevar$ ($N_p$ patches per image). Using $\zstatebb{\statevar}$ as the query and $\Fpatch{1-\statevar}$ (the patch tokens of the other state) as both keys and values, CSA produces a refined summary $\ztilde{\statevar}$ for each state:
\begin{equation}
  \ztilde{\statevar} = \mathrm{CSA}\bigl(Q\!=\!\zstatebb{\statevar}, \ K\!=\!\Fpatch{1-\statevar}, \ V\!=\!\Fpatch{1-\statevar}\bigr), \qquad \text{where} \ \ \statevar\in\stateset.
  \label{eq:csa}
\end{equation}

\paragrapht{Invariant/variant decoders.}
The joint head uses a DPT-style~\citep{ranftl2021dpt} decoder with two branches that share fusion stages: a $9$-channel invariant branch for the joint-wise invariant properties (\emph{e.g.}, type, axis, pivot) and the $2$-channel variant branch for the state-dependent $\theta$ and $d$.
The conditioning vectors are derived from the CSA outputs to match the symmetry of each branch:
\begin{equation*}
\cinv \!=\! (\ztilde{0}+\ztilde{1})/2, \qquad \cvar{\statevar} \!=\! \ztilde{\statevar}, \quad \text{where} \ \ \statevar\in\stateset.
\end{equation*}
Here, the invariant vector $\cinv$ is obtained by symmetrically pooling the two CSA outputs, making it suitable for predicting quantities shared across both states. The variant vector $\cvar{\statevar}$ keeps the state-specific component for predictions that should differ between states.

Let $\mathbf{f}\!\in\!\R^{C\times h\times w}$ denote the per-pixel feature map at a given fusion stage of the decoder.
We first inject geometry by adding the output of a pointwise MLP, which projects the $\Pmap_{i,\statevar}$ into the feature dimension, then apply FiLM~\citep{perez2018film}:
\begin{equation}
\mathbf{f}' = \mathbf{f} + \mathrm{MLP}(\Pmap_{i,\statevar}),
\qquad
\mathbf{f}'' = \boldsymbol{\gamma}\odot \mathbf{f}' + \boldsymbol{\beta},
\label{eq:film}
\end{equation}
where $\boldsymbol{\gamma},\boldsymbol{\beta}$ are linear projections of the branch's conditioning vector ($\cinv$, $\cvar{\statevar}$). The modulated feature $\mathbf{f}''$ is then decoded by each branch into $9$-channel invariant and $2$-channel variant predictions, which are concatenated into the $11$-channel joint map $\Jmap_{i,\statevar}$.

\subsection{Articulation transform}
\label{sec:scgs}

To move Gaussians of the same part together, a part needs to be assigned to each pixel.
At training, we supervise the model to group Gaussians by part for the articulation transform, detailed below, using the true part label, available in the training data.
The joint map is learned end-to-end, without any clustering.
At inference, we assign part pseudo-labels by clustering Gaussians within each joint type using HDBSCAN~\citep{campello2013hdbscan}, a density-based algorithm that infers the number of clusters from the data.
For revolute joints, we cluster on the Plücker line coordinates $(\jaxis,\jaxis\!\times\!\jpivot)$, which uniquely identify the rotation axis as a 3D line.
For prismatic joints, we cluster on the axis direction $\jaxis$ alone, since the pivot is not defined for a translation.
This yields a per-Gaussian pseudo-label for each part $p_n$.

Then, for each part $p$, we obtain a part-level axis $\jaxis_p$, pivot $\jpivot_p$, and reference angle $\bar{\theta}_p$ (revolute) or displacement $\bar{d}_p$ (prismatic) by averaging the per-Gaussian predictions over all Gaussians belonging to the part $p$.
Given target articulation angles $\theta_p^{*}$ or displacements $d_p^*$, we compute the part-level relative motion 
$\Delta\theta_p = \theta_p^{*} - \bar{\theta}_p$ or $\Delta d_p = d_p^{*} - \bar{d}_p$, and apply it rigidly to every Gaussian for the part:
\begin{equation}
  (\boldsymbol{\mu}_n',\,\mathbf{q}_n') =
  \begin{cases}
    \Bigl(R(\jaxis_{p_n},\Delta\theta_{p_n})(\boldsymbol{\mu}_n - \jpivot_{p_n}) + \jpivot_{p_n},\;\; q(\jaxis_{p_n},\Delta\theta_{p_n})\otimes \mathbf{q}_n\Bigr) & \text{revolute,} \\[2pt]
    \bigl(\boldsymbol{\mu}_n + \Delta d_{p_n}\,\jaxis_{p_n},\;\; \mathbf{q}_n\bigr) & \text{prismatic,}
  \end{cases}
  \label{eq:articulation}
\end{equation}
where the rotation matrix $R(\jaxis,\Delta\theta)$ is defined by Rodrigues' formula~\citep{hartley2003multiple}, and the corresponding unit quaternion $q(\jaxis,\Delta\theta)$ represents the same rotation of angle $\Delta\theta$ about axis $\jaxis$, as follows:
\begin{align}
  R(\jaxis,\Delta\theta) &= I + \sin(\Delta\theta)\,[\jaxis]_\times + (1-\cos(\Delta\theta))\,[\jaxis]_\times^2, \\
  q(\jaxis,\Delta\theta) &= \bigl(\cos(\Delta\theta/2),\;\sin(\Delta\theta/2)\,\jaxis\bigr),
\end{align}
where $[\jaxis]_\times$ is the skew-symmetric cross-product matrix of $\jaxis$, and 
$\otimes$ denotes quaternion multiplication.
Applying this transform to every Gaussian yields a state-conditioned Gaussian set $\Gset^{(\statevar^{*})}$ that places each part at its target articulation and can be rendered from any viewpoint.

\subsection{Training}
\label{sec:training}

It is unstable to train the joint head and the rendering objective together from scratch, since an inaccurate joint map often produces incorrectly transformed Gaussians, yielding a misleading photometric signal that, in turn, degrades joint prediction.
We therefore split the training into two stages.
\emph{Stage~1} optimizes the geometry and joint parameters under direct supervision, without any rendering objective.
More details on this supervision and the construction of ground-truth annotations are deferred to Appendix~\cref{app:gt_joint_map}.
\emph{Stage~2} resumes from the \emph{Stage~1}, 
adding a rendering loss (RGB MSE and LPIPS), which both refines appearance quality and end-to-end correction to the joint map through the differentiable articulation transform in \cref{eq:articulation}.
Training in this order is empirically more stable than co-optimizing all losses from scratch.

In Stage 1, we minimize the following loss:
\begin{equation}
  \Lloss{stage1} = \lambda_{\mathrm{p}}\Lloss{pose} + \lambda_{\mathrm{d}}\Lloss{depth} + \lambda_{\mathrm{j}}\Lloss{joint} + \lambda_{\mathrm{c}}\Lloss{consist} + \lambda_{\mathrm{s}}\Lloss{smooth}.
\label{eq:stage1_loss}
\end{equation}
$\Lloss{pose}$ and $\Lloss{depth}$ supervise the camera and depth heads following VGGT~\citep{wang2025vggt}. The remaining three terms supervise the joint map. $\Lloss{joint}$ aggregates per-channel supervision on pixels by
\begin{equation}
  \Lloss{joint} = \mathrm{CE}(\jtype,\jtype^{*}) + \|\jaxis - \jaxis^{*}\|_1 + d_{\perp}(\jpivot, \jpivot^{*};\,\jaxis^{*}) + \mathcal{H}(\theta,\theta^{*}) + \mathcal{H}(d,d^{*}),
\label{eq:l_joint}
\end{equation}
where $\jtype, \jaxis, \jpivot, \theta, d$ are the predicted joint type, axis, pivot, angle, and displacement, starred quantities denote ground truth, and $\mathrm{CE}$ is the cross-entropy loss.
Note that $d_\perp$ is the perpendicular distance from the predicted pivot $\jpivot$ to the GT axis line passing through $\jpivot^{*}$ in direction $\jaxis^{*}$:
\begin{equation}
  d_\perp(\jpivot, \jpivot^{*}; \jaxis^{*}) = \bigl\|(\jpivot - \jpivot^{*}) - \bigl((\jpivot - \jpivot^{*}) \cdot \jaxis^{*}\bigr)\jaxis^{*}\bigr\|_2,
  \label{eq:d_perp}
\end{equation}
which is applied to penalize only deviations from the ground-truth axis line while ignoring offsets along the axis direction. 
$\mathcal{H}$ denotes the Huber loss~\citep{huber1964huber}, applied to $\theta$ and $d$ only for revolute and prismatic pixels, respectively.
Construction of the true joint map is detailed in Appendix~\ref{app:gt_joint_map}.
$\Lloss{consist}$ enforces the invariant prediction to be same across the two states, averaged per part.
Denoting the set of moving parts by $\mathcal{P}$ and the set of pixels belonging to part $p$ in state $\statevar$ by $\Omega_p^{(\statevar)}$, we predict the per-part mean as follows:
\begin{equation}
  \mu_p^{(\statevar)} \;=\; \frac{1}{|\Omega_p^{(\statevar)}|}\sum_{i \in \Omega_p^{(\statevar)}} \Jinv_i.
  \label{eq:consist_mean}
\end{equation}
Then, the consistency loss is given by
\begin{equation}
  \Lloss{consist} \;=\; \frac{1}{|\mathcal{P}|}\sum_{p \in \mathcal{P}}\bigl\|\,\mu_p^{(0)} - \mu_p^{(1)}\,\bigr\|_2.
  \label{eq:l_consist}
\end{equation}
$\Lloss{smooth}$ is a standard part-aware total variation~\citep{rudin1992tvloss} that penalizes $\ell_1$ differences only between adjacent pixels of the same part, preserving part boundaries.

In Stage 2, we add a rendering loss $\Lloss{rgb}$ to $\Lloss{stage1}$:
\begin{equation}
  \Lloss{stage2} = \Lloss{stage1} + \lambda_{\emph{rgb}}\Lloss{rgb},
  \label{eq:stage2_loss}
\end{equation}
where $\Lloss{rgb}$ combines a per-pixel MSE term and a VGG-based LPIPS perceptual term between rendered and ground-truth images.

%% file: section/4_experiments.tex
\section{Experiments}
\label{sec:exp}


\subsection{Experimental settings}
\paragrapht{Dataset.}
We train on PartNet-Mobility~\cite{chang2015shapenet, Mo_2019_CVPR, Xiang_2020_SAPIEN}, a large-scale dataset of URDF-annotated articulated objects across diverse categories, with revolute and prismatic joints connecting movable parts to a static base.
We leverage this dataset to render multi-view, multi-state RGB images for training; full rendering details are provided in Appendix~\ref{app:training_data}.
For evaluation, we hold out a total of $68$ objects from training, consisting of $18$ single-joint and $50$ multi-joint objects.

\paragrapht{Evaluation protocol.}
All methods receive the same $8$ input views ($2$ states $\times$ $4$ views) and are evaluated on $12$ held-out target views per state.
We report results separately on the single-joint and multi-joint splits. Detailed view sampling definitions are provided in Appendix~\ref{app:eval_setup}.
 
\paragrapht{Baselines.}
We compare with baselines from per-object optimization methods built on implicit representations~\cite{liu2023paris,weng2024dta} and 3DGS~\cite{liu2025artgs,kim2025screwsplat,wu2025reartgs}, covering a range of input priors from RGB-only to RGB-D with the number of joints supervision.
 
\paragrapht{Evaluation metrics.}
Following ScrewSplat~\cite{kim2025screwsplat}, we adopt three categories of evaluation metrics: geometry, motion, and appearance.
For \textit{geometry}, we report Chamfer Distance separately for the static part (CD-s), the movable parts 
(CD-m, averaged over all movable parts), and the entire object (CD-w).
For \textit{motion}, we measure the angular error (Ang$_m$, $^\circ$) between the estimated and ground-truth axes, and the axis position error (Pos$_m$) for revolute joints, computed as the minimum distance between corresponding joint axes.
For \textit{appearance}, we report Peak Signal-to-Noise Ratio (PSNR) and Structural Similarity Index Measure (SSIM) on rendered images from the held-out target views.

\paragrapht{Architecture.}
The backbone is the 1B-parameter VGGT aggregator initialized from AnySplat~\citep{jiang2025anysplat} pretrained weights.
We adapt it with rank-16 LoRA~\citep{hu2022lora} adapters (scaling factor $\alpha=16$) on every attention projection, and add a learnable state token of dimension $D=2048$; all other VGGT parameters are frozen.
CSA is a single block with $16$ attention heads and a feature dimension of $2048$;
The Joint DPT head aggregates intermediate features from layers $\{4,11,17,23\}$ of the backbone and progressively fuses them into 
$256$-channel feature maps, from which it produces an $11$-channel output split into the invariant ($9$) and variant ($2$) branches.
The Gaussian head shares the same DPT layout and outputs the AnySplat Gaussian parameters together with a confidence channel.

\paragrapht{Optimization.}
We train with AdamW ($\beta_1\!=\!0.9,\beta_2\!=\!0.95$, weight decay $0.05$) using a learning rate of $1\!\times\!10^{-4}$, with a $1{,}000$-step warm-up followed by cosine decay to $1\!\times\!10^{-5}$. LoRA adapters inside the backbone use $0.1\times$ the base learning rate. Loss weights are $\lambda_{\mathrm{p}}{=}3$, $\lambda_{\mathrm{d}}{=}3$, $\lambda_{\mathrm{j}}{=}5$, $\lambda_{\mathrm{c}}{=}0.5$, $\lambda_{\mathrm{s}}{=}0.1$, and $\lambda_{\mathrm{rgb}}{=}1$, where $\Lloss{rgb} = \mathrm{MSE} + 0.1 \cdot \mathrm{LPIPS}$. Inputs are $448{\times}448$ images, giving $32{\times}32$ patches with the VGGT patch size of $14$. Each batch contains $V\!\cdot\!S\!=\!8$ images from a single object. Stage~1 and Stage~2 run for $80$k and $40$k iterations, respectively, using gradient clipping at $0.5$ and bf16 mixed precision on $4$ NVIDIA RTX A6000 GPUs. The full training schedule takes $\sim\!4$ days.

\paragrapht{Inference.}
At inference, we run a single forward pass on the $8$ input images and form the canonical set $\Gset$ by merging the per-pixel Gaussians with AnySplat's differentiable voxelization at voxel size $0.003$.
We then recover pseudo-labels for the parts by HDBSCAN on Plücker line coordinates 
(\cref{sec:scgs}) and apply the articulation transform of \cref{eq:articulation} to obtain the state-conditioned set $\Gset^{(\statevar^{*})}$, which is rendered with the standard 3DGS rasterizer~\citep{kerbl20233dgs}.


\input{tables/main_results}

\subsection{Main results}

\cref{tab:main_results} compares our method with baselines on both single-joint and multi-joint subsets of PartNet-Mobility, grouped by the prior each method requires.
Among all methods, only ArtSplat and ScrewSplat operate from RGB alone;
the remaining baselines additionally require depth maps, ground-truth joint counts, or both.

\paragrapht{Geometry.}
Among RGB-only methods, ArtSplat achieves the best Chamfer Distance on all three metrics and both splits, consistently outperforming ScrewSplat.
In particular, ArtSplat attains the lowest CD-m overall, including against baselines that rely on depth or joint-count priors.
Since CD-m measures how well movable parts are reconstructed, it directly reflects whether the model correctly identifies which regions move and how they transform---the core capability ArtSplat is designed to learn.
Notably, ArtSplat's geometry remains robust as the number of joints increases, which we attribute to cross-state attention explicitly comparing the two articulation states.
 
\paragrapht{Motion.}
Accurate axis estimation is critical, as even small angular errors compound into large displacements for parts far from the joint origin.
Existing methods struggle to recover accurate joint axis orientation, with most reporting angular errors above $30^\circ$ regardless of the input prior.
In contrast, ArtSplat predicts joint axes with substantially lower angular error on both single- and multi-joint splits, demonstrating robust axis estimation that does not degrade for objects with more joints. 
 
\paragrapht{Appearance.}
ArtSplat achieves competitive appearance quality on both splits, which is noteworthy given the fundamental efficiency gap: per-object optimization methods fit each scene for hundreds of iterations over sampled views, whereas ArtSplat produces all Gaussians in a single feed-forward pass from only 8 input images.

\begin{figure}[t]
    \centering
    \includegraphics[width=1\linewidth]{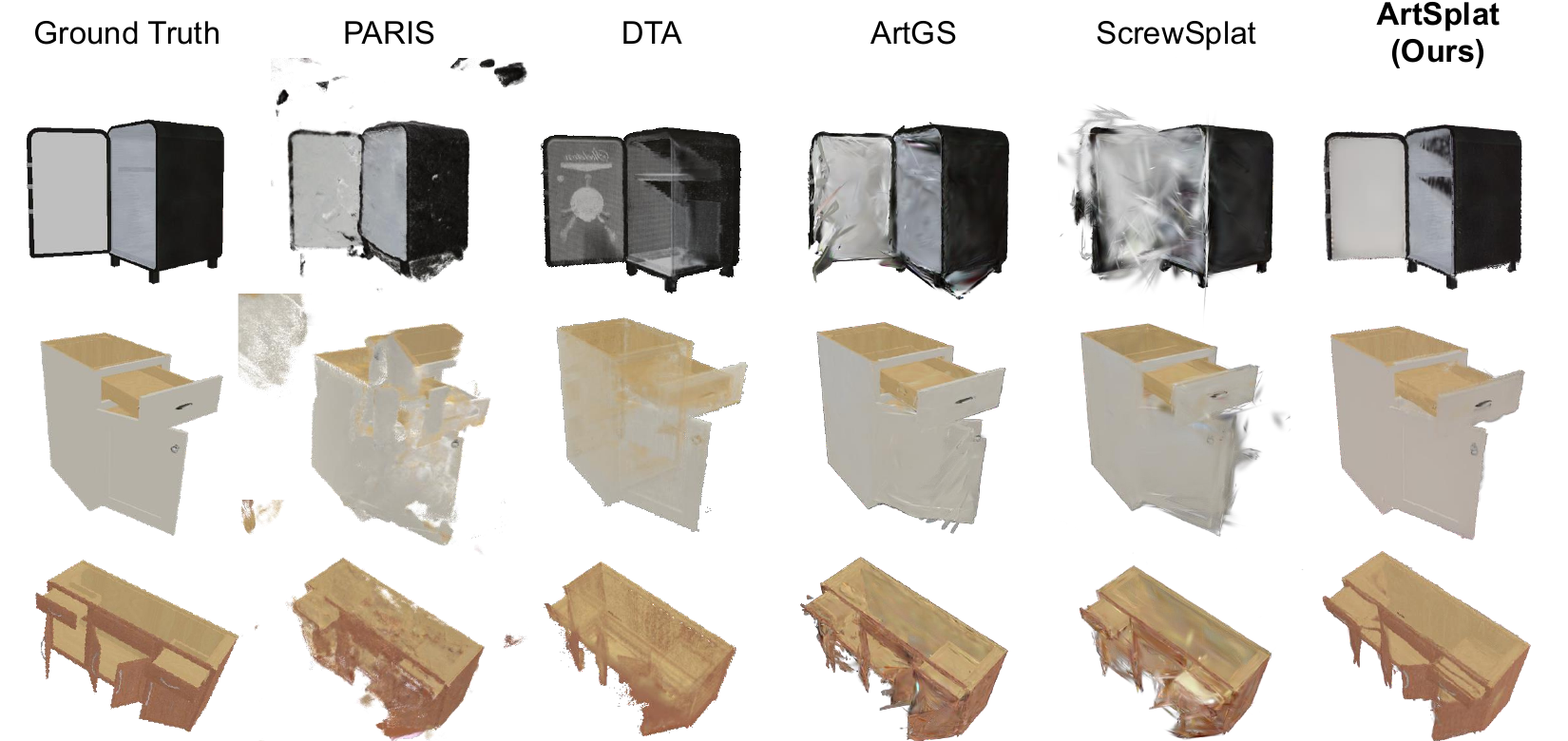}
        \caption{Qualitative comparison of novel-view renderings via Gaussian rasterization. Baselines exhibit ghosting and misaligned edges around the joints due to inaccurate axis estimation, whereas ArtSplat produces clean renderings of both static and movable parts.}
    \label{fig:qualitative_comparison_rendering}
\end{figure}

\begin{figure}[t]
    \centering
    \includegraphics[width=1\linewidth]{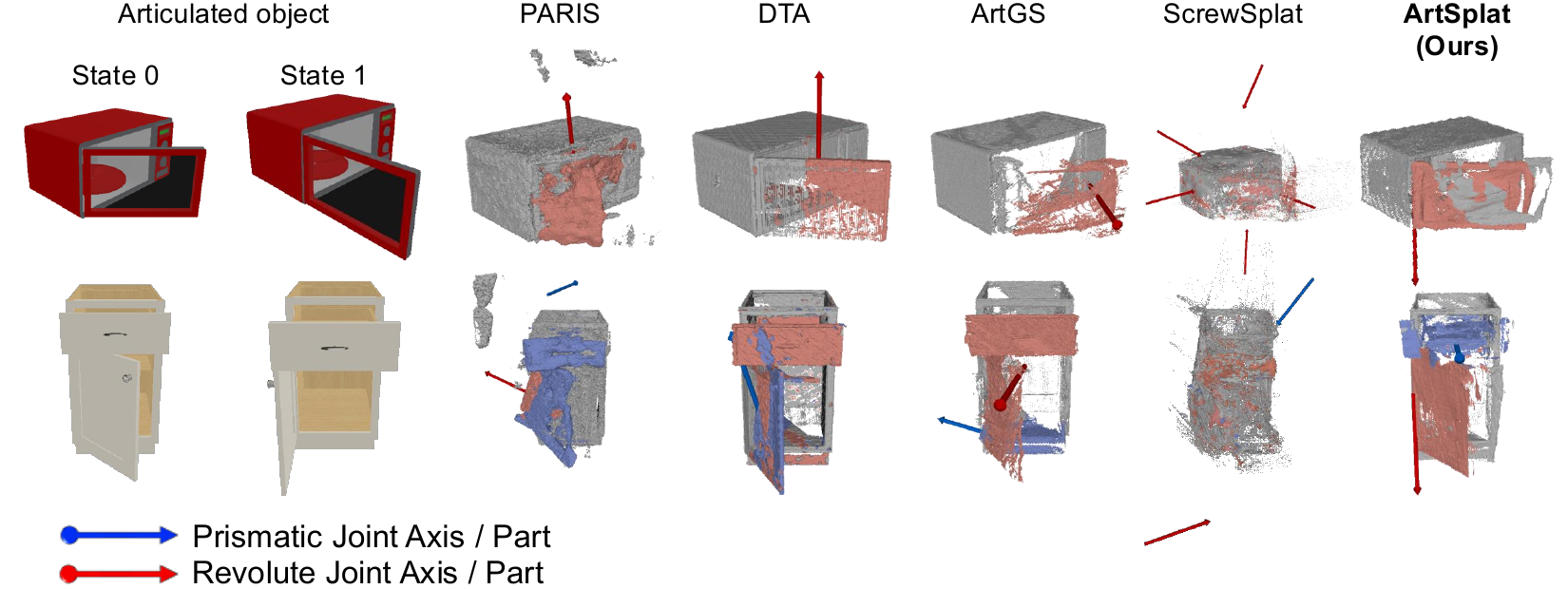}
    \caption{Qualitative comparison of extracted meshes and predicted joint axes.}
    \label{fig:qualitative_comparison_mesh}
\end{figure}

\begin{figure}[h!]
    \centering
    \includegraphics[width=0.8\linewidth]{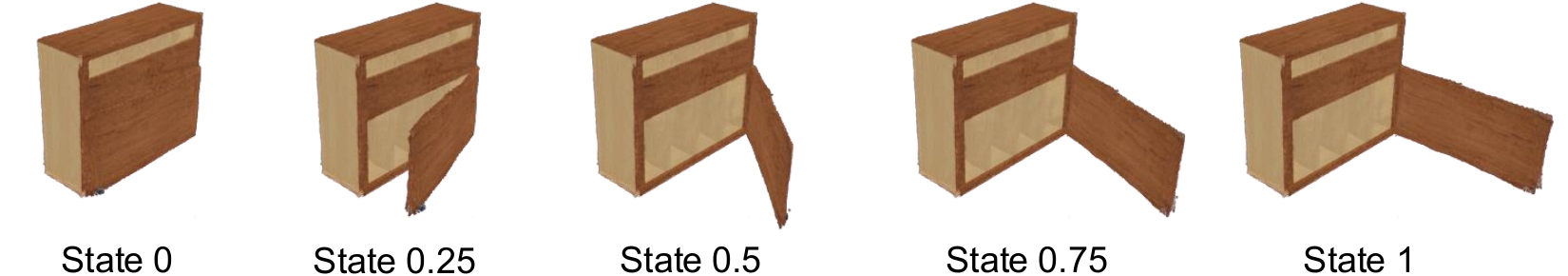}
    \caption{Illustration of Novel state rendering. Given multi-view RGB observations, ArtSplat reconstructs state-conditioned Gaussians and renders the object at novel articulation states $s \in [0, 1]$, where $s{=}0$ and $s{=}1$ denote the fully closed and fully open configurations, respectively.}
    \label{fig:novel_state}
\end{figure}

\paragrapht{Qualitative comparison.}
As shown in \cref{fig:qualitative_comparison_rendering}, baselines often mispredict joint axes, causing movable parts to be rendered at incorrect positions or orientations and producing visible artifacts.
In contrast, ArtSplat predicts joint axes accurately, yielding clean renderings of movable parts.
\cref{fig:qualitative_comparison_mesh} shows that ArtSplat also extracts competitive meshes while localizing joint axes more accurately than the baselines.
As further shown in \cref{fig:novel_state}, ArtSplat can render the object at arbitrary articulation states $s\!\in\![0,1]$, smoothly interpolating between fully closed ($s\!=\!0$) and fully open ($s\!=\!1$) configurations.

\input{tables/inference_speed}

\paragrapht{Inference speed.}

\cref{tab:inference_speed} compares the inference speed of competing methods.
As a feed-forward model, ArtSplat reconstructs each object in under 2 seconds, achieving over 400$\times$ speedup over existing baselines and approximately 700$\times$ over ScrewSplat, which also uses only RGB inputs.


\input{tables/ablation}

\subsection{Ablation studies}
\label{sec:ablation}

\paragrapht{Cross-state Attention.}
We ablate the cross-state attention by disabling it while retaining the state token, so the state token cannot attend to features from the other state but still provides FiLM conditioning to the joint head.
As in \cref{tab:ablation_main}, this substantially degrades axis and angle estimation accuracy while leaving per-state geometry largely intact.
This asymmetric degradation indicates that axis and angle estimation relies mostly on the state token aggregating features across states.

\paragrapht{State Token Conditioning.}
We further remove the state token in addition to cross-state attention, so image tokens are passed directly to the DPT joint head without any state-specific signal.
As reported in \cref{tab:ablation_main}, this variant performs worst across all joint metrics, with a particularly large drop relative to the cross-state-attention-only ablation.
The gap indicates that the state token and cross-state attention contribute beyond what either component alone does.

\paragrapht{Invariant/variant DPT.}
We ablate the invariant/variant split by replacing the dual-branch DPT head with a single shared head that jointly regresses all eleven channels, without FiLM conditioning.
As seen in \cref{tab:ablation_main}, the single-head variant achieves comparable axis estimation but consistently underperforms on angle and displacement.
This asymmetric degradation shows that making the invariant/variant distinction structural rather than learned yields more accurate variant predictions without sacrificing invariant-attribute quality.

\paragrapht{Two-stage training.}
We compare our two-stage training with a single-stage variant that simultaneously optimizes all losses, including the rendering loss $\mathcal{L}_{\text{rgb}}$, from the first iteration.
To ensure a fair comparison, both schedules use the same total number of training iterations.
The last row of \cref{tab:ablation_main} indicates that the performance significantly drops compared to our two-stage training.
We attribute this to potentially unstable gradients through the articulated transformation that corrupt the joint and point heads early in training, due to the loss $\mathcal{L}_{\text{rgb}}$ backpropagated through Gaussians parameterized by still-noisy joint predictions.
This leads to oscillating joint losses and eventually degraded performance.
Our two-stage schedule avoids this issue by introducing photometric supervision only after the joint and geometry predictions become sufficiently stable, allowing $\mathcal{L}_{\text{rgb}}$ to act as a refinement signal rather than a source of optimization noise.
It also improves efficiency, since our Stage~1 omits differentiable rasterization and photometric supervision, making each iteration substantially cheaper.

%% file: tables/main_results.tex
\begin{table*}[t]
    \centering
    \caption{Quantitative comparison on the PartNet-Mobility dataset. 
    Best results are in \textbf{bold}, and the second best are \underline{underlined}.}
    \label{tab:main_results}
    \resizebox{\textwidth}{!}{%
        \begin{tabular}{l l l | ccc cc cc}
        \toprule
        \textbf{Split} & \textbf{Method} & \textbf{Prior}
        & \multicolumn{3}{c}{Geometry ($\downarrow$)}
        & \multicolumn{2}{c}{Motion ($\downarrow$)}
        & \multicolumn{2}{c}{Appearance ($\uparrow$)} \\
        \cmidrule(lr){4-6}
        \cmidrule(lr){7-8}
        \cmidrule(lr){9-10}
        & & & CD-w & CD-s & CD-m & Ang$_m$ & Pos$_m$ & PSNR & SSIM \\
        \midrule

        \multirow{6}{*}{\rotatebox[origin=c]{90}{Single-joint}}

            & PARIS~\cite{liu2023paris}
            & RGB + number of joints
            & 0.022
            & 0.029
            & 0.294
            & 55.96
            & \phantom{0}\textbf{0.166}
            & 20.51
            & 0.909 \\

            & DTA~\cite{weng2024dta}
            & RGB-D + number of joints
            & \textbf{0.004}
            & \textbf{0.008}
            & 0.145
            & \underline{33.79}
            & \phantom{0}2.305
            & 20.23
            & 0.822 \\

            & ArtGS~\cite{liu2025artgs}
            & RGB-D + number of joints
            & \underline{0.013}
            & \underline{0.010}
            & \underline{0.102}
            & 64.67
            & \phantom{0}1.165
            & \underline{28.01}
            & \underline{0.924} \\

            & ReArtGS~\cite{wu2025reartgs}
            & RGB + number of joints
            & 0.045
            & 0.129
            & 0.443
            & 47.87
            & \phantom{0}1.993
            & 23.44
            & 0.865 \\

            & ScrewSplat~\cite{kim2025screwsplat}
            & RGB
            & 0.036
            & 0.039
            & 0.301
            & 26.27
            & \phantom{0}2.401
            & 22.93
            & 0.918 \\

            & ArtSplat (Ours)
            & RGB
            & 0.031
            & 0.023
            & \textbf{0.028}
            & \phantom{0}\textbf{6.65}
            & \phantom{0}\underline{0.436}
            & \textbf{28.70}
            & \textbf{0.941} \\

        \midrule

        \multirow{5}{*}{\rotatebox[origin=c]{90}{Multi-joint}}

            & PARIS~\cite{liu2023paris}
            & RGB + number of joints
            & \textbf{0.017}
            & \underline{0.029}
            & \underline{0.335}
            & 53.08
            & \phantom{0}\textbf{0.221}
            & 20.03
            & \underline{0.913} \\

            & DTA~\cite{weng2024dta}
            & RGB-D + number of joints
            & \underline{0.025}
            & 0.053
            & 0.527
            & \underline{44.48}
            & 22.790
            & 19.56
            & 0.813 \\

            & ArtGS~\cite{liu2025artgs}
            & RGB-D + number of joints
            & 0.096
            & 0.102
            & 0.782
            & 49.31
            & \phantom{0}1.825
            & \underline{25.21}
            & 0.893 \\


            & ScrewSplat~\cite{kim2025screwsplat}
            & RGB
            & 0.035
            & 0.044
            & 0.405
            & 46.78
            & \phantom{0}3.368
            & 23.49
            & 0.912 \\

            & ArtSplat (Ours)
            & RGB
            & 0.034
            & \textbf{0.024}
            & \textbf{0.030}
            & \phantom{0}\textbf{5.79}
            & \phantom{0}\underline{0.864}
            & \textbf{26.30}
            & \textbf{0.921} \\

        \bottomrule
        \end{tabular}%
    }
\end{table*}

%% file: tables/inference_speed.tex
\begin{table*}[t]
\centering
\small
\setlength{\tabcolsep}{5pt}
\caption{Per-object inference speed on PartNet-Mobility. We compare average wall-clock time to reconstruct a single articulated object, measured on a single NVIDIA RTX A6000 GPU with 8 input views.}
\label{tab:inference_speed}
\begin{tabular}{l l l c}
\toprule
\textbf{Type} & \textbf{Method} & \textbf{Prior} & \textbf{Time per object $\downarrow$} \\
\midrule
\multirow{5}{*}{\emph{per-object optimization}}
    & PARIS~\cite{liu2023paris}            & RGB + number of joints       & \underline{12m 57s} \\
    & DTA~\cite{weng2024dta}               & RGB-D + number of joints     & 56m 07s \\
    & ArtGS~\cite{liu2025artgs}            & RGB-D + number of joints     & 15m 50s \\
    & REArtGS~\cite{wu2025reartgs}         & RGB + number of joints       & 1h 37m \\
    & ScrewSplat~\cite{kim2025screwsplat}  & RGB                          & 23m 05s \\
\midrule
\emph{feed-forward} & \textbf{ArtSplat (Ours)} & RGB & \textbf{$<$2s} \\
\bottomrule
\end{tabular}
\end{table*}

%% file: tables/ablation.tex
\begin{table*}[t]
\centering
\small
\setlength{\tabcolsep}{3pt}
\caption{Architecture and training-schedule ablations. Each row removes or replaces one component and is retrained with the same schedule as the full model.}
\label{tab:ablation_main}
\begin{tabular}{l ccc cc cc}
\toprule
& \multicolumn{3}{c}{Geometry ($\downarrow$)}
& \multicolumn{2}{c}{Motion ($\downarrow$)}
& \multicolumn{2}{c}{Appearance ($\uparrow$)} \\
\cmidrule(lr){2-4} \cmidrule(lr){5-6} \cmidrule(lr){7-8}
Method
& CD-w & CD-s & CD-m
& Ang$_m$ & Pos$_m$
& PSNR & SSIM \\
\midrule
\textbf{ArtSplat (full)}                  & \textbf{0.034} & \textbf{0.024} & \textbf{0.030} & \phantom{0}\textbf{5.79}  & \textbf{0.864} & \textbf{26.30} & \textbf{0.921} \\
\midrule
w/o cross-state attention                 & 0.067          & 0.029          & 0.082          & 18.74          & 1.852          & 24.18          & 0.901 \\
w/o state token conditioning              & 0.112          & 0.041          & 0.156          & 31.46          & 2.917          & \underline{22.51}          & 0.879 \\
w/o Inv/Var split (single branch)         & \underline{0.042}          & \underline{0.026}          & \underline{0.041}          & \phantom{0}\underline{7.12}           & 1.643          & 25.07          & \underline{0.913} \\
Single-stage joint training                    & 0.058          & 0.030          & 0.069          & 12.85          & \underline{1.428}          & 23.74          & 0.895 \\
\bottomrule
\end{tabular}
\end{table*}

%% file: section/5_conclusion.tex
\section{Conclusion}
\label{sec:conclusion}
We propose ArtSplat, a feed-forward framework that reconstructs articulated objects and their joint parameters from sparse, uncalibrated RGB views in a single forward pass, eliminating per-object optimization.
A novel per-pixel joint map representation and a joint prediction module that estimates joint parameters from inter-state differences are successfully integrated into the feed-forward reconstruction pipeline.
Extensive experiments on PartNet-Mobility demonstrate that ArtSplat successfully reconstructs both geometry and motion of articulated objects at significantly faster speed than per-object optimization methods.

\paragrapht{Limitations and future work.}
Our model currently handles only single-degree-of-freedom joints connected directly to a static base.
Extending it to kinematic chains, where multiple joints are connected in series as in a robotic arm, and validating on real-world captures would be promising directions for future work.

%% file: section/X_suppl.tex
\appendix

\crefalias{section}{appendix}
\crefalias{subsection}{appendix}
\crefalias{subsubsection}{appendix}
\section*{Appendix}

\setcounter{table}{0}
\setcounter{figure}{0}
\renewcommand{\thetable}{\Roman{table}}
\renewcommand{\thefigure}{\Roman{figure}}

\section{Training data details}
\label{app:training_data}

\subsection{Multi-view rendering}
\label{app:rendering}

For each training object, we pre-render RGB images, depth maps, and
per-part segmentation masks from $48$ camera viewpoints across $16$
articulation states, yielding $48 \times 16 = 768$ frames per object.

\paragrapht{Camera placement.}
Cameras are distributed on the upper hemisphere via stratified jittering, looking at the object center with no roll.
The hemisphere is covered by two layers:
a \emph{main layer} of $5$ elevation bins $\times$ $8$ azimuth bins ($=40$ views)
spanning elevation $[0^{\circ}, 72^{\circ}]$ and full $360^{\circ}$ azimuth,
and a \emph{top layer} of $2$ elevation bins $\times$ $4$ azimuth bins ($=8$ views)
spanning elevation $[72^{\circ}, 85^{\circ}]$.

\paragrapht{Articulation state sampling.}
We sample $16$ states per object using stratified jittering over the normalized articulation range $[0,1]$.
The range is divided into $16$ equal bins, and one articulation value is randomly sampled within each bin.
For multi-joint objects, each joint receives an independent random permutation of the $16$ bins, so joint configurations are decorrelated across states.

\subsection{Training view sampling}
\label{app:training_loading}

At each training iteration, we sample one object and draw $2$ states uniformly
from the $16$ pre-rendered states, then select $4$ views per state from the
$48$ available viewpoints, forming a batch of $V{\cdot}S{=}8$ images.
All supervision signals---ground-truth depth, part segmentation, camera
extrinsics, and the joint map.

\section{Ground-truth joint map details}
\label{app:gt_joint_map}

The ground-truth joint map $\Jmap^{*}_{i,s}\!\in\!\R^{H\times W\times 11}$ is constructed per frame from per-pixel part segmentation labels, joint annotations (type, axis, pivot), and joint positions (angle or displacement) from the PartNet-Mobility dataset.
The channel layout is summarized in \cref{tab:joint_map_channels}.
Following VGGT~\citep{wang2025vggt}, all geometric quantities are expressed in a \emph{canonical frame} defined by the first camera $[R_0 \mid \mathbf{t}_0]$, and further divided by a radius normalization factor $\bar{r}$ (the mean distance of foreground 3D points from the canonical origin) so that the scale-normalized space is shared across the point map, camera translations, and joint map.

For each pixel with part label $\ell$ ($-1$\,=\,background, $0$\,=\,static base, $\geq\!1$\,=\,moving part), the 11 channels are populated as follows:

\begin{itemize}[leftmargin=1em, itemsep=0.1em]

\item \textbf{Channels 0--2} (joint type):
One-hot encoding of the joint type:
static $[1,0,0]$, revolute $[0,1,0]$, or prismatic $[0,0,1]$.

\item \textbf{Channels 3--5} (axis direction):
The joint axis direction associated with the corresponding articulated part, transformed into the canonical frame and normalized to a unit vector.

\item \textbf{Channels 6--8} (pivot position):
The joint pivot position associated with the articulated part, transformed into the canonical frame and normalized by $\bar{r}$.

\item \textbf{Channel 9} (rotation angle $\theta$):
For revolute joints, the articulation value is converted into a rotation angle using the joint motion range.

\item \textbf{Channel 10} (displacement $d$):
For prismatic joints, the articulation value is converted into a linear displacement using the joint motion range and normalized by $\bar{r}$.

\end{itemize}

All pixels belonging to the same moving part share identical values for channels 0--8 (invariant properties), while channels 9--10 (variant properties) vary according to the articulation state of the current frame.

\input{tables/jointmap}

\section{Evaluation setup}
\label{app:eval_setup}

\subsection{Evaluation view sampling}
All cameras are placed on the upper hemisphere looking at the object center.
Views are generated via \emph{stratified jittering}: we partition
the elevation range into $N_e$ equal intervals and the azimuth range into $N_a$
equal intervals, forming an $N_e \times N_a$ grid of bins.  Within each bin,
one camera is placed at a uniformly random elevation and azimuth, yielding
$N_e \cdot N_a$ views in total.

\paragrapht{Input views.}
Each method receives $8$ images ($2$ states $\times$ $4$ views).
The $4$ views per state use $N_e{=}1, N_a{=}4$ over elevation
$[5^{\circ}, 60^{\circ}]$ and azimuth $[-180^{\circ}, 180^{\circ})$.

\paragrapht{Target views.}
Each state is evaluated on $12$ held-out views, using $N_e{=}3, N_a{=}4$
over elevation $[0^{\circ}, 90^{\circ}]$ and azimuth
$[-180^{\circ}, 180^{\circ})$.
The wider elevation range ensures that target views include viewpoints
above and below the input distribution, testing generalization rather than
interpolation alone.

\section{Additional qualitative results}
\label{app:qual}

\begin{figure}
    \centering
    \includegraphics[width=1\linewidth]{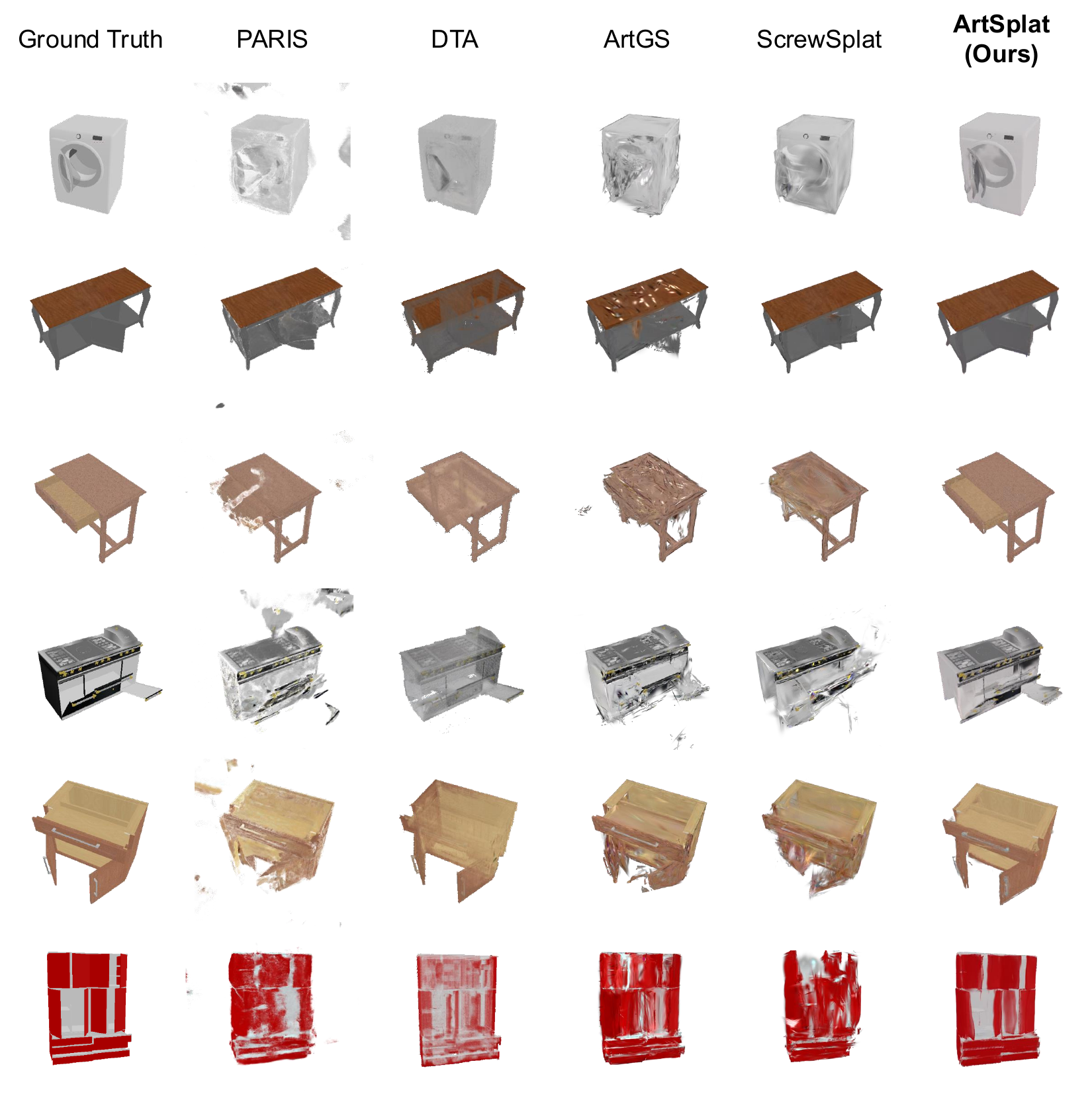}
    \caption{Additional qualitative comparisons on novel-view rendering across diverse articulated objects from the test set.}
    \label{fig:add_qual}
\end{figure}

\cref{fig:add_qual} provides additional qualitative comparisons on three single-joint and three multi-joint objects from the test set.
Notably, the bottom example contains 8 revolute and 6 prismatic joints, and ArtSplat still achieves favorable rendering quality compared to the baselines, highlighting its scalability to complex articulation structures.

\begin{figure}[h]
    \centering
    \includegraphics[width=0.80\linewidth]{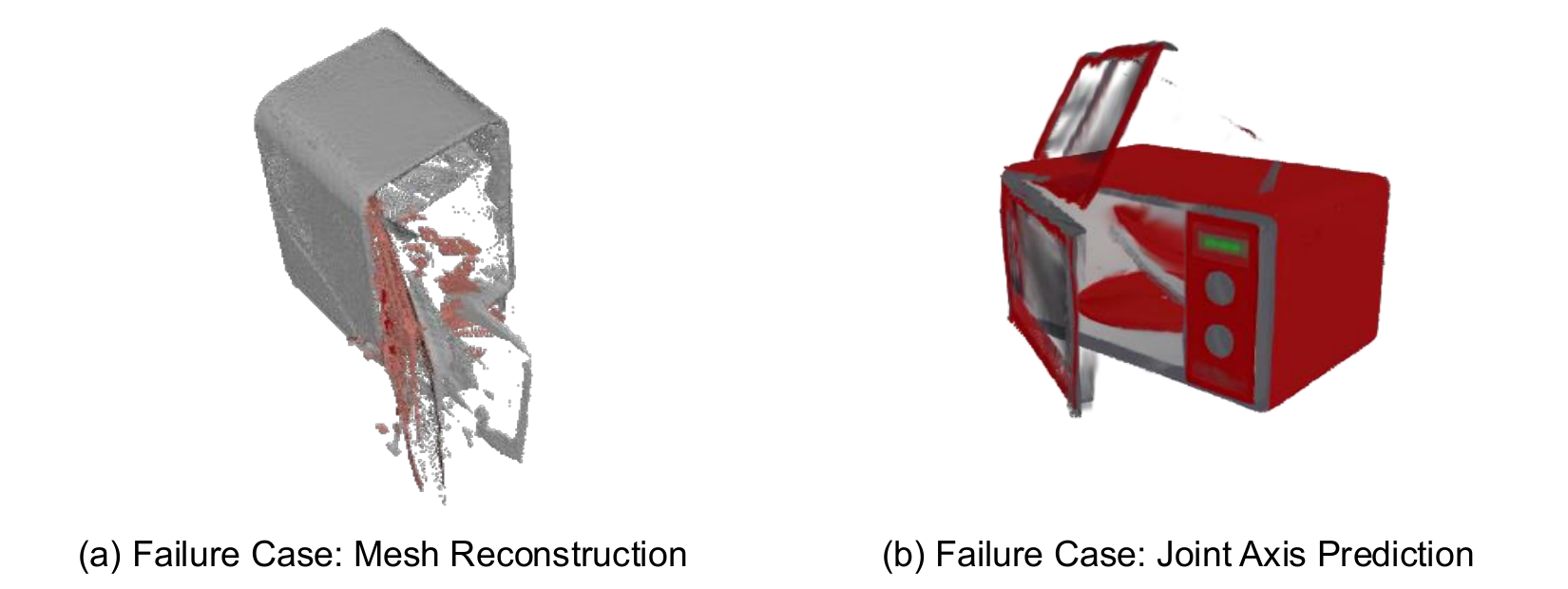}
    \caption{Representative failure cases.
    \emph{Left:} Mesh reconstruction artifacts.
    \emph{Right:} Rendering errors caused by inaccurate joint axis prediction
    when the articulation difference between two states is small.}
    \label{fig:failure_cases}
\end{figure}

\paragrapht{Failure cases.}
\cref{fig:failure_cases} illustrates two representative failure cases of ArtSplat. First, 3D Gaussians are inherently unstructured primitives without explicit surface connectivity, which makes high-fidelity mesh reconstruction challenging.
Adopting surface-aligned representations such as 2D Gaussian Splatting~\cite{huang20242dgs} is a promising direction to mitigate this limitation.
Second, when the articulation change between the two observed states is small, the predicted joint axis can deviate significantly from the ground truth, leading to incorrect articulation transforms and consequently degraded novel-view rendering.
Leveraging external priors, \emph{e.g.}, vision-language models that explicitly reason about inter-state differences, could help resolve such ambiguities.

%% file: tables/jointmap.tex
\begin{table}[h]
\centering
\small
\caption{Channel layout of the per-pixel joint map $\Jmap$ ($11$ channels). Invariant channels describe the joint identity and must agree across states; variant channels describe state-dependent motion.}
\vspace{0.5em}
\label{tab:joint_map_channels}
\begin{tabular}{cllc}
\toprule
\textbf{Channel} & \textbf{Quantity} & \textbf{Type} & \textbf{Invariant} \\
\midrule
0--2  & joint type logits  & static / revolute / prismatic & $\checkmark$ \\
3--5  & axis direction $\jaxis$ & unit 3-vector            & $\checkmark$ \\
6--8  & pivot $\jpivot$    & 3-D world point              & $\checkmark$ \\
9     & rotation $\theta$  & radian (revolute)            & per-state \\
10    & displacement $d$   & meter (prismatic)            & per-state \\
\bottomrule
\end{tabular}
\end{table}

%% file: main.bib
@String(AAAI = "Proceedings of the AAAI Conference on Artificial Intelligence (AAAI)")

@String(CVPR = "Proceedings of the IEEE/CVF Conference on Computer Vision and Pattern Recognition (CVPR)")

@String(ECCV = "Proceedings of the European Conference on Computer Vision (ECCV)")

@String(ICCV = "Proceedings of the IEEE/CVF International Conference on Computer Vision (ICCV)")

@String(ICLR = "Proceedings of the International Conference on Learning Representations (ICLR)")

@String(ICRA = "Proceedings of the IEEE International Conference on Robotics and Automation (ICRA)")

@String(NeurIPS = "Advances in Neural Information Processing Systems (NeurIPS)")

@String(PAKDD = "Proceedings of the Pacific-Asia Conference on Knowledge Discovery and Data Mining (PAKDD)")

@String(CoRL = "Proceedings of the Conference on Robot Learning (CoRL)")

@String(IC3DV = "Proceedings of the International Conference on 3D Vision (3DV)")

@String(TOG = "ACM Transactions on Graphics (TOG)")

@inproceedings{liu2023paris,
  title={{PARIS: Part-level Reconstruction and Motion Analysis for Articulated Objects}},
  author={Liu, Jiayi and Mahdavi-Amiri, Ali and Savva, Manolis},
  booktitle=ICCV,
  year={2023}
}

@inproceedings{weng2024dta,
  title={{Neural Implicit Representation for Building Digital Twins of Unknown Articulated Objects}},
  author={Weng, Yijia and Wen, Bowen and Tremblay, Jonathan and Blukis, Valts and Fox, Dieter and Guibas, Leonidas and Birchfield, Stan},
  booktitle=CVPR,
  year={2024}
}

@inproceedings{jiang2022ditto,
  title={{Ditto: Building Digital Twins of Articulated Objects from Interaction}},
  author={Jiang, Zhenyu and Hsu, Cheng-Chun and Zhu, Yuke},
  booktitle=CVPR,
  year={2022}
}

@inproceedings{tseng2022cla,
  title={{CLA-NeRF: Category-Level Articulated Neural Radiance Field}},
  author={Tseng, Wei-Cheng and Liao, Hung-Ju and Yen-Chen, Lin and Sun, Min},
  booktitle=ICRA,
  year={2022},
}

@inproceedings{guo2025articulatedgs,
  title={{ArticulatedGS: Self-supervised Digital Twin Modeling of Articulated Objects using 3D Gaussian Splatting}},
  author={Guo, Junfu and Xin, Yu and Liu, Gaoyi and Xu, Kai and Liu, Ligang and Hu, Ruizhen},
  booktitle=CVPR,
  year={2025}
}

@inproceedings{liu2025artgs,
  title={{ArtGS: Building Interactable Replicas of Complex Articulated Objects via Gaussian Splatting}},
  author={Liu, Yu and Jia, Baoxiong and Lu, Ruijie and Ni, Junfeng and Zhu, Song-Chun and Huang, Siyuan},
  booktitle=ICLR,
  year={2025}
}

@inproceedings{wu2025reartgs,
  title={{REArtGS: Reconstructing and Generating Articulated Objects via 3D Gaussian Splatting with Geometric and Motion Constraints}},
  author={Wu, Di and Liu, Liu and Linli, Zhou and Huang, Anran and Song, Liangtu and Yu, Qiaojun and Wu, Qi and Lu, Cewu},
  booktitle=NeurIPS,
  year={2025}
}

@inproceedings{kim2025screwsplat,
  title={{ScrewSplat: An End-to-End Method for Articulated Object Recognition}},
  author={Kim, Seungyeon and Ha, Junsu and Kim, Young Hun and Lee, Yonghyeon and Park, Frank C},
  booktitle=CoRL,
  year={2025}
}

@inproceedings{lin2025splart,
  title={{SplArt: Articulation Estimation and Part-Level Reconstruction with 3D Gaussian Splatting}},
  author={Lin, Shengjie and Fang, Jiading and Irshad, Muhammad Zubair and Guizilini, Vitor Campagnolo and Ambrus, Rares Andrei and Shakhnarovich, Greg and Walter, Matthew R},
  booktitle=ICCV,
  year={2025}
}

@article{yu2025part^2,
  title={{Part$^2$GS: Part-aware Modeling of Articulated Objects using 3D Gaussian Splatting}},
  author={Yu, Tianjiao and Shah, Vedant and Wahed, Muntasir and Shen, Ying and Nguyen, Kiet A and Lourentzou, Ismini},
  journal={arXiv:2506.17212},
  year={2025}
}

@inproceedings{shen2025gaussianart,
  title={{GaussianArt: Unified Modeling of Geometry and Motion for Articulated Objects}},
  author={Shen, Licheng and Zhang, Saining and Li, Honghan and Yang, Peilin and Huang, Zihao and Zhang, Zongzheng and Zhao, Hao},
  booktitle=IC3DV,
  year={2025}
}

@inproceedings{charatan2024pixelsplat,
  title={{pixelSplat: 3D Gaussian Splats from Image Pairs for Scalable Generalizable 3D Reconstruction}},
  author={Charatan, David and Li, Sizhe Lester and Tagliasacchi, Andrea and Sitzmann, Vincent},
  booktitle=CVPR,
  year={2024}
}

@inproceedings{chen2024mvsplat,
  title={{MVSplat: Efficient 3D Gaussian Splatting from Sparse Multi-View Images}},
  author={Chen, Yuedong and Xu, Haofei and Zheng, Chuanxia and Zhuang, Bohan and Pollefeys, Marc and Geiger, Andreas and Cham, Tat-Jen and Cai, Jianfei},
  booktitle=ECCV,
  year={2024},
}

@inproceedings{xu2025depthsplat,
  title={{DepthSplat: Connecting Gaussian Splatting and Depth}},
  author={Xu, Haofei and Peng, Songyou and Wang, Fangjinhua and Blum, Hermann and Barath, Daniel and Geiger, Andreas and Pollefeys, Marc},
  booktitle=CVPR,
  year={2025}
}

@inproceedings{ye2025nopose,
  title={{No Pose, No Problem: Surprisingly Simple 3D Gaussian Splats from Sparse Unposed Images}},
  author={Ye, Botao and Liu, Sifei and Xu, Haofei and Li, Xueting and Pollefeys, Marc and Yang, Ming-Hsuan and Peng, Songyou},
  booktitle=ICLR,
  year={2025}
}

@article{smart2024splatt3r,
  title={{Splatt3R: Zero-shot Gaussian Splatting from Uncalibrated Image Pairs}},
  author={Smart, Brandon and Zheng, Chuanxia and Laina, Iro and Prisacariu, Victor Adrian},
  journal={arXiv:2408.13912},
  year={2024}
}

@article{jiang2025anysplat,
  title={{AnySplat: Feed-forward 3D Gaussian Splatting from Unconstrained Views}},
  author={Jiang, Lihan and Mao, Yucheng and Xu, Linning and Lu, Tao and Ren, Kerui and Jin, Yichen and Xu, Xudong and Yu, Mulin and Pang, Jiangmiao and Zhao, Feng and others},
  journal=TOG,
  volume={44},
  number={6},
  pages={1--16},
  year={2025},
  publisher={ACM New York, NY, USA}
}

@inproceedings{wang2024dust3r,
  title={{DUSt3R: Geometric 3D Vision Made Easy}},
  author={Wang, Shuzhe and Leroy, Vincent and Cabon, Yohann and Chidlovskii, Boris and Revaud, Jerome},
  booktitle=CVPR,
  year={2024}
}

@inproceedings{leroy2024grounding,
  title={{Grounding Image Matching in 3D with MASt3R}},
  author={Leroy, Vincent and Cabon, Yohann and Revaud, J{\'e}r{\^o}me},
  booktitle=ECCV,
  year={2024},
}

@inproceedings{wang2025vggt,
  title={{VGGT: Visual Geometry Grounded Transformer}},
  author={Wang, Jianyuan and Chen, Minghao and Karaev, Nikita and Vedaldi, Andrea and Rupprecht, Christian and Novotny, David},
  booktitle=CVPR,
  year={2025}
}

@inproceedings{yang2025fast3r,
  title={{Fast3R: Towards 3D Reconstruction of 1000+ Images in One Forward Pass}},
  author={Yang, Jianing and Sax, Alexander and Liang, Kevin J and Henaff, Mikael and Tang, Hao and Cao, Ang and Chai, Joyce and Meier, Franziska and Feiszli, Matt},
  booktitle=CVPR,
  year={2025}
}

@inproceedings{zhang2025flare,
  title={{FLARE: Feed-forward Geometry, Appearance and Camera Estimation from Uncalibrated Sparse Views}},
  author={Zhang, Shangzhan and Wang, Jianyuan and Xu, Yinghao and Xue, Nan and Rupprecht, Christian and Zhou, Xiaowei and Shen, Yujun and Wetzstein, Gordon},
  booktitle=CVPR,
  year={2025}
}

@inproceedings{zhang2024monst3r,
  title={{MonST3R: A Simple Approach for Estimating Geometry in the Presence of Motion}},
  author={Zhang, Junyi and Herrmann, Charles and Hur, Junhwa and Jampani, Varun and Darrell, Trevor and Cole, Forrester and Sun, Deqing and Yang, Ming-Hsuan},
  booktitle=ICLR,
  year={2025}
}

@inproceedings{wang2025continuous,
  title={{Continuous 3D Perception Model with Persistent State}},
  author={Wang, Qianqian and Zhang, Yifei and Holynski, Aleksander and Efros, Alexei A and Kanazawa, Angjoo},
  booktitle=CVPR,
  year={2025}
}

@inproceedings{chen2025easi3r,
  title={{Easi3R: Estimating Disentangled Motion from DUSt3R Without Training}},
  author={Chen, Xingyu and Chen, Yue and Xiu, Yuliang and Geiger, Andreas and Chen, Anpei},
  booktitle=ICCV,
  year={2025}
}

@inproceedings{li2026art,
  title={{ART: Articulated Reconstruction Transformer}},
  author={Li, Zizhang and Zhang, Cheng and Li, Zhengqin and Howard-Jenkins, Henry and Lv, Zhaoyang and Geng, Chen and Wu, Jiajun and Newcombe, Richard and Engel, Jakob and Dong, Zhao},
  booktitle=CVPR,
  year={2026}
}

@inproceedings{yuan2025larm,
  title={{LARM: A Large Articulated Object Reconstruction Model}},
  author={Yuan, Sylvia and Shi, Ruoxi and Wei, Xinyue and Zhang, Xiaoshuai and Su, Hao and Liu, Minghua},
  booktitle={Proceedings of the SIGGRAPH Asia Conference Papers (SIGGRAPH Asia)},
  year={2025}
}

@InProceedings{Xiang_2020_SAPIEN,
author = {Xiang, Fanbo and Qin, Yuzhe and Mo, Kaichun and Xia, Yikuan and Zhu, Hao and Liu, Fangchen and Liu, Minghua and Jiang, Hanxiao and Yuan, Yifu and Wang, He and Yi, Li and Chang, Angel X. and Guibas, Leonidas J. and Su, Hao},
title = {{SAPIEN}: A SimulAted Part-based Interactive ENvironment},
booktitle =CVPR,
year = {2020}}

@InProceedings{Mo_2019_CVPR,
author = {Mo, Kaichun and Zhu, Shilin and Chang, Angel X. and Yi, Li and Tripathi, Subarna and Guibas, Leonidas J. and Su, Hao},
title = {{PartNet}: A Large-Scale Benchmark for Fine-Grained and Hierarchical Part-Level {3D} Object Understanding},
booktitle =CVPR,
year = {2019}
}

@article{chang2015shapenet,
title={{Shapenet: An information-rich 3d model repository}},
author={Chang, Angel X and Funkhouser, Thomas and Guibas, Leonidas and Hanrahan, Pat and Huang, Qixing and Li, Zimo and Savarese, Silvio and Savva, Manolis and Song, Shuran and Su, Hao and others},
journal={arXiv:1512.03012},
year={2015}
}

@inproceedings{ranftl2021dpt,
  title={{Vision Transformers for Dense Prediction}},
  author={Ranftl, Ren{\'e} and Bochkovskiy, Alexey and Koltun, Vladlen},
  booktitle=CVPR,
  year={2021}
}

@inproceedings{perez2018film,
  title={{FiLM: Visual Reasoning with a General Conditioning Layer}},
  author={Perez, Ethan and Strub, Florian and De Vries, Harm and Dumoulin, Vincent and Courville, Aaron},
  booktitle=AAAI,
  year={2018}
}

@inproceedings{hu2022lora,
  title={{LoRA: Low-Rank Adaptation of Large Language Models}},
  author={Hu, Edward J and Shen, Yelong and Wallis, Phillip and Allen-Zhu, Zeyuan and Li, Yuanzhi and Wang, Shean and Wang, Liang and Chen, Weizhu and others},
  booktitle=ICLR,
  year={2022}
}

@inproceedings{campello2013hdbscan,
  title={{Density-Based Clustering Based on Hierarchical Density Estimates}},
  author={Campello, Ricardo JGB and Moulavi, Davoud and Sander, J{\"o}rg},
  booktitle=PAKDD,
  year={2013},
}

@article{mildenhall2021nerf,
  title={{NeRF}: Representing {S}cenes as {N}eural {R}adiance {F}ields for {V}iew {S}ynthesis},
  author={Mildenhall, Ben and Srinivasan, Pratul P and Tancik, Matthew and Barron, Jonathan T and Ramamoorthi, Ravi and Ng, Ren},
  journal={Communications of the ACM},
  volume={65},
  number={1},
  pages={99--106},
  year={2021},
  publisher={ACM New York, NY, USA}
}

@article{kerbl20233dgs,
  title={3D Gaussian Splatting for Real-Time Radiance Field Rendering.},
  author={Kerbl, Bernhard and Kopanas, Georgios and Leimk{\"u}hler, Thomas and Drettakis, George},
  journal={ACM Trans. Graph.},
  volume={42},
  number={4},
  pages={139--1},
  year={2023}
}

@article{oquab2023dinov2,
  title={{DINOv2: Learning Robust Visual Features without Supervision}},
  author={Oquab, Maxime and Darcet, Timoth{\'e}e and Moutakanni, Th{\'e}o and Vo, Huy and Szafraniec, Marc and Khalidov, Vasil and Fernandez, Pierre and Haziza, Daniel and Massa, Francisco and El-Nouby, Alaaeldin and others},
  journal={arXiv:2304.07193},
  year={2023}
}

@article{rudin1992tvloss,
  title={Nonlinear total variation based noise removal algorithms},
  author={Rudin, Leonid I and Osher, Stanley and Fatemi, Emad},
  journal={Physica D: nonlinear phenomena},
  volume={60},
  number={1-4},
  pages={259--268},
  year={1992},
  publisher={Elsevier}
}

@article{huber1964huber,
    author = {Peter J. Huber},
    title = {{Robust Estimation of a Location Parameter}},
    volume = {35},
    journal = {The Annals of Mathematical Statistics},
    number = {1},
    publisher = {Institute of Mathematical Statistics},
    pages = {73 -- 101},
    year = {1964},
}

@book{hartley2003multiple,
  title={{Multiple View Geometry in Computer Vision}},
  author={Hartley, Richard and Zisserman, Andrew},
  year={2003},
  publisher={Cambridge university press}
}

@inproceedings{huang20242dgs,
  title={{2D Gaussian Splatting for Geometrically Accurate Radiance Fields}},
  author={Huang, Binbin and Yu, Zehao and Chen, Anpei and Geiger, Andreas and Gao, Shenghua},
  booktitle={ACM SIGGRAPH 2024 conference papers},
  pages={1--11},
  year={2024}
}
